%% file: acl_crver.tex
\pdfoutput=1

\documentclass[11pt]{article}
\usepackage[whole]{bxcjkjatype}
\usepackage{acl}

\usepackage{times}
\usepackage{latexsym}

\usepackage[T1]{fontenc}
\usepackage[whole]{bxcjkjatype}

\usepackage[utf8]{inputenc}

\usepackage{microtype}

\usepackage{lingmacros}
\usepackage{amsmath,amssymb}
\usepackage{mathtools}
\usepackage{ascmac}
\usepackage{proof,lscape,caption}
\usepackage{multirow}
\usepackage{color}
\usepackage{stmaryrd}
\newcommand{\bhline}[1]{\noalign{\hrule height #1}}
\newcommand{\rulelabelsize}{\scriptsize}
\newcommand{\FA}{\mbox{\rulelabelsize $<$}}
\newcommand{\BA}{\mbox{\rulelabelsize $>$}}

\newcommand{\SR}[1]{\begin{tabular}{c}$#1$\end{tabular}}

\title{Compositional Semantics and Inference System \\for Temporal Order based on Japanese CCG}

\author{Tomoki Sugimoto \and Hitomi Yanaka \\
        The University of Tokyo\\
        \texttt{\{sugimoto.tomoki, hyanaka\}@is.s.u-tokyo.ac.jp}}

\begin{document}
\maketitle
\begin{abstract}
Natural Language Inference (NLI) is the task of determining whether a premise entails a hypothesis.
NLI with temporal order is a challenging task because tense and aspect are complex linguistic phenomena involving interactions with temporal adverbs and temporal connectives.
To tackle this, temporal and aspectual inference has been analyzed in various ways in the field of formal semantics.
However, a Japanese NLI system for temporal order based on the analysis of formal semantics has not been sufficiently developed.
We present a logic-based NLI system that considers temporal order in Japanese based on compositional semantics via Combinatory Categorial Grammar (CCG) syntactic analysis.
Our system performs inference involving temporal order by using axioms for temporal relations and automated theorem provers.
We evaluate our system by experimenting with Japanese NLI datasets that involve temporal order.
We show that our system outperforms previous logic-based systems as well as current deep learning-based models.
\end{abstract}

\section{Introduction}
Natural Language Inference (NLI) is the task of determining whether a premise entails a hypothesis.
In particular, NLI involving temporal expressions is crucial.
(\ex{1}) is an example of English NLI involving temporal expressions.

\enumsentence{\textit{P}: I arrived in April 2021.\\$\overline{\textit{H}\text{: I arrived before May 2021. (entailment)}}$}

\noindent The inference example with temporal expressions is challenging.
This is because we need to represent the meaning of sentences that contain temporal adverbs like \textit{before} and \textit{in}, temporal expressions like \textit{April} 2021, and verb tenses like \textit{arrived}, and to compute temporal order of events written in the sentences.

\citet{thukral-etal-2021-probing} showed that deep learning-based models~\cite{liu2019roberta, he2020deberta} trained on a standard NLI dataset such as Multi-Genre Natural Language Inference (MultiNLI; \citet{williams-etal-2018-broad}) failed to perform simple temporal inference as in (\ex{0}).
Furthermore, deep learning-based models have performed poorly on challenging NLI datasets that involve various temporal inferences such as FraCaS~\cite{cooper1996} for English and JSeM~\cite{kawazoe2015inference} for Japanese.

Recently, logical inference systems based on compositional semantics~\cite{bos-markert-2005-recognising,abzianidze-2015-tableau,mineshima-etal-2015-higher,mineshima-etal-2016-building,bernardy-chatzikyriakidis-2017-type,bernardy2020fracas,Onishi2020ccg} (i.e., semantics in which the meaning of a phrase is determined compositionally from the syntax and the meaning of the lexicon contained in the phrase) achieved high accuracy in FraCaS and JSeM.
However, most previous systems did not cover temporal inference.

In addition, because most previous research on NLI has focused on English, research on other languages is desirable.
In particular, research on NLI in Japanese is still in its infancy and is limited to deep learning-based systems using pre-trained language models and a few logical inference systems~\cite{mineshima-etal-2016-building,Onishi2020ccg}.
\citet{Onishi2020ccg} attempted to implement a Japanese logical inference system for temporal inference.
However, the focus of this previous research was limited to a few temporal clauses in Japanese, and temporal adverbs are out of scope.
Thus, there is still room for improvement in the accuracy of temporal inference in Japanese.

In this study, our aim is to realize the compositional semantics and a logical inference system for temporal inference in Japanese based on Combinatory Categorial Grammar (CCG)~\cite{steedman2000syntactic, Bekki2010} to derive a transparent syntax-semantics interface and the analysis of tense and aspect studied in formal semantics~\cite{Kamp1993-KAMFDT,yoshimoto2000tense,kaufmann2011temporal,Utsugi2015towards,ogihara2017tense,jacobsen_2018}.
We focus on temporal order and develop a Japanese logical inference system for temporal order.

In our system, a CCG parser first parses the premise and hypothesis sentences and converts them into CCG trees.
Based on the analysis of the compositional semantics, we then modify the obtained CCG trees.
Next, using ccg2lambda~\cite{martinez-gomez-etal-2016-ccg2lambda}, the meaning of the whole sentence is derived as a logical form.
Finally, we attempt to prove the entailment relations between the obtained logical forms by an automated theorem prover Vampire~\cite{kovacs2013first}.

We experiment with two NLI datasets involving temporal order in Japanese: JSeM and a Japanese translation of the NLI dataset focusing on temporal inference~\cite{thukral-etal-2021-probing}.
We compare our system with the previous Japanese logical inference system~\cite{Onishi2020ccg} and the Japanese BERT model~\cite{devlin-etal-2019-bert}.
Our experiments show that our system outperforms previous logical inference systems as well as current deep learning-based models.
Our system will be available for research use at \url{https://github.com/ynklab/ccgtemp}.

\section{Background}\label{sec:background}
Tense and aspect are important linguistic phenomena related to temporal expressions.
This section provides standard background on the semantics of temporal expressions in Japanese, which have been analyzed in previous studies~\cite{yoshimoto2000tense,kaufmann2011temporal,Utsugi2015towards,ogihara2017tense,jacobsen_2018}.

In Japanese, verb tense is classified into past (\textit{-ta}) and non-past (\textit{-ru}), and aspect is classified into stative (like \textit{iru}) and non-stative (like \textit{kuru}).
The temporal interpretation of a matrix clause (i.e., a clause that contains a subordinate clause) is determined by the combination of tense and aspect, and is expressed by the constraints imposed on the relation between speech time and reference time.
Speech time represents the time that a sentence is uttered, and reference time is a concept proposed by \citet{Reichenbach1947-REIEOS} and refers to the time used with location time (i.e., time when an event occurs) and speech time to represent the meaning of tense.
Table~\ref{tab:tense_aspect_relation} shows the temporal interpretation of a matrix clause determined by the combination of tense and aspect and example sentences corresponding to each combination.

\input{sources/table232_1_tense_aspect_relation.tex}

To analyze the temporal interpretation of embedded clauses, the concepts of absolute tense and relative tense are necessary.
Absolute tense means that the temporal interpretation is determined by the relation between the speech time and the reference time, as in the matrix clause.
However, relative tense means an interpretation in which the temporal interpretation does not depend on the relation between the speech time and the reference time.
We explain the details with examples in Section~\ref{compsemclause}.

This paper uses CCG to formalize the syntactic analysis of our method and analyzes the compositional semantics of temporal expressions based on the analysis by \citet{kaufmann2011temporal}.

\section{Compositional Semantics and Inference for Tense}
\subsection{Semantic Representations for Verb Tense}\label{subsec:compsemanticsverb}

\input{sources/tree311_1_example_derivation.tex}

This section explains the semantic representations for verb tense.
Consider the following sentences.
\eenumsentence{
    \item\shortex{2}
        {Taro-ga & kuru}
        {Taro-\textsc{nom} & come-\textsc{np}}
        {‘Taro is coming’}
    \item\shortex{2}
        {Taro-ga & kita}
        {Taro-\textsc{nom} & come-\textsc{p}}
        {‘Taro came’}
}
(\ex{0}a) is non-past tense (\textsc{np}), and (\ex{0}b) is past tense (\textsc{p}).
(\ex{0}a) means that the event of Taro's coming occurs after the speech time, whereas (\ex{0}b) means that the event
occurred before the speech time.
Thus, for the speech time $s$ and the reference time $r$, $r > s$ in (\ex{0}a) and $r < s$ in (\ex{0}b).
Here, $r$ and $s$ both represent intervals and $r < s$ means the end of the interval $r$ is before the beginning of the interval $s$.
Another interpretation of time is instance semantics, which treats time as an instance, but in this study, we follow the standard treatment of time as an interval~\cite{Kamp1993-KAMFDT,bernardy2020fracas}.

\begin{figure}[h]
    \centering
    \includegraphics[width=7cm]{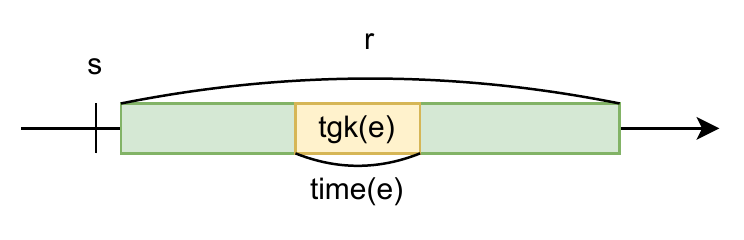}
    \vspace{-3mm}
    \caption{Temporal interpretation of (\ex{-1}a)}
    \label{fig:temporalinterpretation1}
\end{figure}

\begin{figure}[h]
    \centering
    \includegraphics[width=7cm]{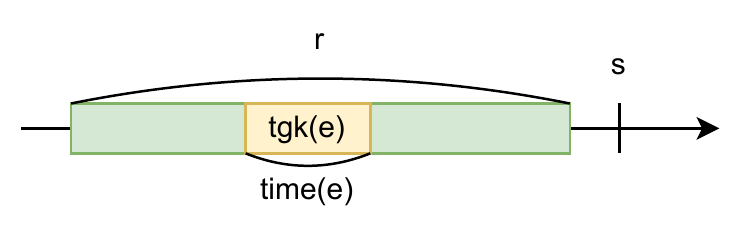}
    \vspace{-3mm}
    \caption{Temporal interpretation of (\ex{-1}b)}
    \label{fig:temporalinterpretation2}
\end{figure}

Following \citet{Kamp1993-KAMFDT}, in this study, the time of an event is represented by its relationship with the reference time.
Then, the meaning of (\ex{0}a) and (\ex{0}b) can be expressed by the following logical expressions, where $\mathsf{tgk}$ is the predicate that represents the event Taro's coming,
$\mathsf{time}$ is the function that returns the time when the event occurred and $e$ is a variable representing the event.
\eenumsentence{
    \item $\exists e.(\mathsf{tgk}(e) \wedge \mathsf{time}(e) \subseteq r \wedge r > s)$
    \item $\exists e.(\mathsf{tgk}(e) \wedge \mathsf{time}(e) \subseteq r \wedge r < s)$
}
The meanings of (\ex{0}a) and (\ex{0}b) are as shown in the Figure~\ref{fig:temporalinterpretation1} and Figure~\ref{fig:temporalinterpretation2}.
Figure \ref{fig:example_derivation} shows the CCG derivation tree for (\ex{-1}a).

\vspace{-7mm}

\subsection{Semantic Representations for Temporal Clause}\label{compsemclause}
Next, consider the following sentences with an embedded clause.
\eenumsentence{
    \item\shortex{4}
        {Taro-ga & kuru & mae-ni & oyoida}
        {Taro-\textsc{nom} & come-\textsc{np} & before-\textsc{loc} & swim-\textsc{p}}
        {‘I swam before Taro came’}
    \item\shortex{4}
        {Taro-ga & kita & ato-ni & oyoida}
        {Taro-\textsc{nom} & come-\textsc{np} & after-\textsc{loc} & swim-\textsc{p}}
        {‘I swam after Taro came’}
}
In (\ex{0}a), the embedded clause is the non-past tense, and in (\ex{0}b), the embedded clause is the past tense.
As mentioned in Section~\ref{sec:background}, the temporal meaning of embedded clauses is interpreted using ``relative tense.''
Thus, the temporal meaning of embedded clauses is determined not by the relation between the speech time and the reference time of the embedded clause but by the relation between the reference time of the matrix clause and the reference time of the embedded clause.
For the reference time of the embedded clause $t$ and the reference time of the matrix clause $r$, we then have $t > r$ in (\ex{0}a), and $t < r$ in (\ex{0}b).

\begin{figure}[h]
    \centering
    \includegraphics[width=7cm]{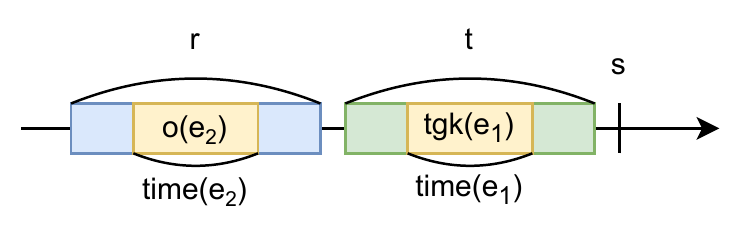}
    \vspace{-3mm}
    \caption{Temporal interpretation of (\ex{-2}a)}
    \label{fig:temporalinterpretation3}
\end{figure}

\begin{figure}[h]
    \centering
    \includegraphics[width=7cm]{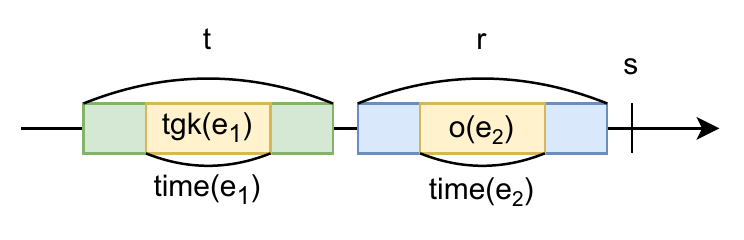}
    \vspace{-3mm}
    \caption{Temporal interpretation of (\ex{-2}b)}
    \label{fig:temporalinterpretation4}
\end{figure}

Therefore, using the same predicates and functions as Section~\ref{subsec:compsemanticsverb}, the meaning of the embedded clauses can be expressed by the following logical formulas.
\eenumsentence{
    \item $\exists e.(\mathsf{tgk}(e) \wedge \mathsf{time}(e) \subseteq t \wedge t > r)$
    \item $\exists e.(\mathsf{tgk}(e) \wedge \mathsf{time}(e) \subseteq t \wedge t < r)$
}
By combining these logical formulas with the meanings of the matrix clauses interpreted in the same way as Section~\ref{subsec:compsemanticsverb}, the meanings of sentences with the embedded clauses can be expressed by the following logical formulas, where $\mathsf{o}$ is the predicate that represents the event of my swimming.
\eenumsentence{
    \item $\exists t. (\exists e_1. (\mathsf{tgk}(e_1) \wedge \mathsf{time}(e_1) \subseteq t \wedge t > r) \wedge \exists e_2. (\mathsf{o}(e_2) \wedge \mathsf{time}(e_2) \subseteq r \wedge r < s))$
    \item $\exists t. (\exists e_1. (\mathsf{tgk}(e_1) \wedge \mathsf{time}(e_1) \subseteq t \wedge t < r) \wedge \exists e_2. (\mathsf{o}(e_2) \wedge \mathsf{time}(e_2) \subseteq r \wedge r < s))$
}
The meanings of (\ex{0}a) and (\ex{0}b) are as shown in the Figure~\ref{fig:temporalinterpretation3} and Figure~\ref{fig:temporalinterpretation4}.

\noindent This study interprets the temporal meaning of sentences with embedded clauses in this way.

\subsection{Semantic Representations for Temporal Adverb}
\subsubsection{Syntactic analysis}\label{subsubsec:ccgtree}
An example of the temporal adverbs targeted in this paper is shown in bold in the following.
\enumsentence{\shortex{7}
{Taro-ga & \textbf{4} & \textbf{gatsu} & \textbf{3} & \textbf{nichi} & \textbf{izen-ni} & kita}
{Taro-\textsc{nom} & \textbf{4} & \textbf{month} & \textbf{3} & \textbf{day} & \textbf{before} & come-\textsc{p}}
{‘Taro came \textbf{before April 3}’}
}

\noindent More generally, we analyze temporal adverbs comprising various types of absolute temporal expressions (e.g., date, day of the week, and time) and temporal connectives \textit{izen} (\textit{before}) and \textit{ikou} (\textit{after}).
Absolute temporal expressions are temporal expressions that do not depend on the speech time, in contrast to relative temporal expressions such as \textit{today} that depend on the speech time.
In this study, temporal adverbs containing relative temporal expressions are out of scope and left for future work.

In temporal adverbs containing absolute temporal expressions, the particle \textit{-ni} is unnecessary.
For example, the following three sentences are all acceptable and have the same meaning.
\eenumsentence{\item \shortex{7}
{4 & gatsu & 3 & nichi & ni & Taro-ga & kita}
{4 & month & 3 & day & on & Taro-\textsc{nom} & come-\textsc{p}}
{‘Taro came on April 3’}
\item \shortex{6}
{4 & gatsu & 3 & nichi, & Taro-ga & kita}
{4 & month & 3 & day & Taro-\textsc{nom} & come-\textsc{p}}
{‘Taro came on April 3’}
\item \shortex{6}
{4 & gatsu & 3 & nichi & Taro-ga & kita}
{4 & month & 3 & day & Taro-\textsc{nom} & come-\textsc{p}}
{‘Taro came on April 3’}
}
Thus, \textit{-ni} can be analyzed as a separation of clauses like a comma and does not have any meaning.
Before considering the syntactic category of \textit{-ni}, let us consider absolute temporal expressions.
As shown in (\ex{0}c), absolute temporal expressions are combined with sentences such as \textit{Taro-ga kita}.
Therefore, $S/S$ is assigned as the syntactic category of the absolute temporal expression 4 \textit{gatsu} 3 \textit{nichi}.
As mentioned above, because \textit{-ni} plays the role of connecting the preceding and following clauses, $(S/S)\backslash(S/S)$ is appropriate as its syntactic category.

In addition,
absolute temporal expressions like 4 \textit{gatsu} 3 \textit{nichi} can be a noun phrase $NP$, as in Figure~\ref{fig:example_ccg_tree}.
In this example, the syntactic category of 4 \textit{gatsu} 3 \textit{nichi} is $NP$, and the syntactic category of \textit{izen} is $(S/S)\backslash NP$.
We explain the reason why absolute temporal expressions are used as both $NP$ and $S/S$ from a semantic perspective in the next paragraph.

\input{sources/tree321_1_example_CCG_tree}

\input{sources/table332_1_semantic_templates.tex}

\subsubsection{Semantic analysis}\label{subsubsec:semantictemplates}
We treat absolute temporal expressions (e.g., 4 \textit{gatsu} 3 \textit{nichi} (\textit{April} 3)) as multi-word expressions.
Consider the expression 4 \textit{gatsu} 3 \textit{nichi}.
We can decompose the expression into four constituents as follows.
\begin{align*}
    [4\ gatu\ 3\ nichi] &= [4\ gatu][3\ nichi] \\
    &= [[4][gatu]][[3][nichi]]
\end{align*}
A current Japanese CCG parser~\cite{yoshikawa-etal-2017-ccg} analyzes each constituent as the syntactic category $4 = NP, gatsu = (NP/NP)\backslash NP, 3 = NP,$ and $nichi = NP/NP$, respectively.
The semantic template for $NP$ is $\lambda E\ N\ F. ^\exists x.(N(E,x) \wedge F(x))$, which means ``some bound variable $x$ is associated with the word $E$.''
Now 4 and 3 are both $NP$, so 4 and 3 have different bound variables associated with them.
This bound variable refers to the interval.
Essentially, because 4 \textit{gatsu} 3 \textit{nichi} refers to only one interval, 4 and 3 need to be associated with the same interval.
The correct meaning cannot be derived when 4 and 3 are associated with different bound variables.

Thus, we treat temporal expressions such as 4 \textit{gatsu} 3 \textit{nichi} as multi-word expressions and set up a semantic template as shown in Table~\ref{tab:semantic_templates}.
This semantic template allows us to derive the meaning of a temporal expression associated with only one bound variable.
In this template, the function $\mathsf{normalized\_time}$ takes interval as an argument and returns its actual time, which can be set in the format YYYYMMDDHH from absolute temporal expressions.
For example, for interval $x$, which represents \textit{April} 3, the value is $\mathsf{normalized\_time}(x)=0000040300$.
In this example, year and hour are not explicitly written, so zero-padding is applied to them.

As shown in Figure~\ref{fig:example_ccg_tree},
4 \textit{gatsu} 3 \textit{nichi} functions as $NP$ when connected to \textit{izen} and as $S/S$ when used by itself.
This phenomenon can be analyzed as follows.
Temporal expressions such as 4 \textit{gatsu} 3 \textit{nichi} and 4 \textit{gatsu} 3 \textit{nichi izen} play the role of representing the time of the sentence.
Consider the following sentences.
\enumsentence{
    \shortex{7}
        {4 & gatsu & 3 & nichi & ni & Taro-ga & kita}
        {4 & month & 3 & day & on & Taro-\textsc{nom} & come-\textsc{p}}
        {‘Taro came on April 3’}
}
\enumsentence{
    \shortex{7}
        {4 & gatsu & 3 & nichi & izen-ni & Taro-ga & kita}
        {4 & month & 3 & day & before  & Taro-\textsc{nom} & come-\textsc{p}}
        {‘Taro came before April 3’}
}
In (\ex{-1}), the location time of the event \textit{Taro-ga kita} (\textit{Taro came}) is 4 \textit{gatsu} 3 \textit{nichi} (\textit{April} 3), and in (\ex{0}), the location time of the event \textit{Taro-ga kita} (\textit{Taro came}) is 4 \textit{gatsu} 3 \textit{nichi izen} (\textit{before April} 3).
The expressions that represent temporal adverbs such as 4 \textit{gatsu} 3 \textit{nichi} (\textit{April} 3) and 4 \textit{gatsu} 3 \textit{nichi izen} (\textit{before April} 3) must have the syntactic category of $S/S$, so 4 \textit{gatsu} 3 \textit{nichi} changes from $NP$ to $S/S$.

Next, the semantic template for \textit{izen} was determined as shown in Table~\ref{tab:semantic_templates}.
The temporal meaning of \textit{izen} is represented as the lambda expression $\lambda x. \mathsf{before}(j3,x)$, which indicates that the expression ``doing before $x$'' means ``doing in $j3$ before $x$.''
Finally, the meaning of temporal expressions can be derived by setting up a template with \textit{-ni} and a comma as meaningless words, as described in Section~\ref{subsubsec:ccgtree}.

\begin{figure*}[t]
    \centering
    \includegraphics[width=14cm]{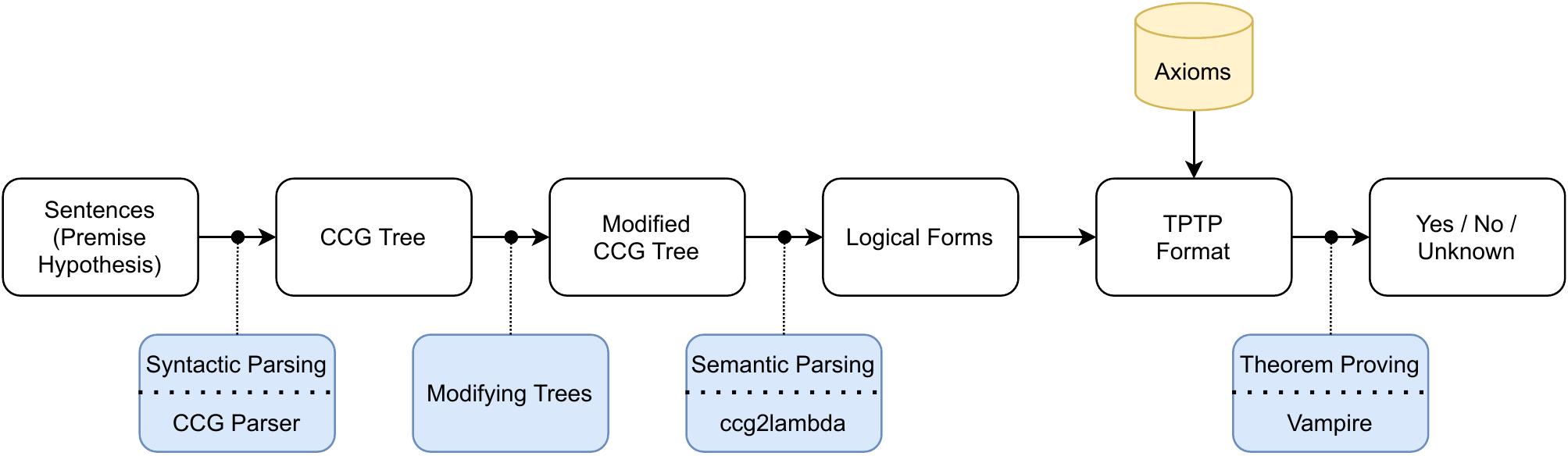}
    \caption{Overview of our system}
    \label{fig:overview}
\end{figure*}

\subsection{Inference with Tense}
We introduce a set of axioms for temporal relations and temporal expressions to perform inference for temporal order.
\citet{allen-1983-maintaining} defined 13 relations between time intervals.
The previous logic-based inference system~\citep{Onishi2020ccg} introduced 169 axioms for these 13 temporal relations.
Six of the 13 temporal relations, $\mathsf{meets}$, $\mathsf{met\_by}$, $\mathsf{starts}$, $\mathsf{started\_by}$, $\mathsf{finishes}$, and $\mathsf{finished\_by}$ are special cases of other relations in implementing axioms.
For example, $\mathsf{meets}$ is a special case of $\mathsf{before}$ where the end of the preceding interval coincides with the beginning of the following interval.
$\mathsf{meets}$ is necessary for inferences involving temporal clauses such as \textit{soon after}.
Thus, we consider that those six relations are redundant in performing the temporal inference involving temporal order in this study.
We therefore merged them into the most similar relations: merged $\mathsf{meets}$ into $\mathsf{before}$, $\mathsf{met\_by}$ into $\mathsf{after}$, $\mathsf{starts}$ into $\mathsf{during}$, $\mathsf{started\_by}$ into $\mathsf{contains}$, $\mathsf{finishes}$ into $\mathsf{during}$, and $\mathsf{finished\_by}$ into $\mathsf{contains}$, respectively.
In summary, we introduce 49 axioms corresponding to seven temporal relations: $\mathsf{before}$, $\mathsf{after}$, $\mathsf{overlaps}$, $\mathsf{overlapped\_by}$, $\mathsf{during}$, $\mathsf{contains}$, and $\mathsf{equal}$.

In addition, we speculate 30 additional axioms for temporal expressions in Japanese such as \textit{izen} (\textit{before}) and \textit{ikou} (\textit{after}), and those for identity conditions of speech times between premises and hypotheses.
Table~\ref{tab:axioms} shows examples of the axioms.

\input{sources/table341_1_axioms.tex}

\section{System Overview}\label{sec:systemoverview}

Figure~\ref{fig:overview} shows the pipeline of our system.
Our system consists of three main steps.
First, natural language sentences of premises and hypotheses are converted into modified CCG trees by CCG parsing and modifying trees.
Next, a meaning from the semantic templates is assigned to each lexical item.
The semantics in lexical items are then composed by ccg2lambda to derive a logical formula that represents the meaning of the whole sentence.
Finally, an automated theorem prover determines whether the logical formula of the hypothesis is provable from the logical formula of the premises.
In this section, we describe each of these steps.

\subsection{Syntactic Analysis}\label{subsec:syntacticanalysis}
The syntactic analysis, which obtains CCG parsing trees of input sentences, consists of two steps.
First, we use the tokenizer to tokenize sentences and a CCG parser to obtain a CCG tree.
We use depccg~\cite{yoshikawa-etal-2017-ccg}, a standard Japanese CCG parser, trained on the Japanese CCGBank~\cite{uematsu-etal-2013-integrating} for the first step.

Second, if the sentence contains temporal expressions, we extract the subtrees in which the leaves are temporal expressions from the CCG tree of the whole sentence.
The extracted CCG subtree is then transformed into an appropriate form.
Figure~\ref{fig:before_modify_tree} and Figure~\ref{fig:after_modify_tree} show the temporal expression subtrees 4 \textit{gatsu} 3 \textit{nichi ni} (\textit{on April} 3) before and after the conversion.
As another possible way of implementation for obtaining correct CCG trees for temporal expressions, we can improve the CCG parser itself.
However, to do that, we need to re-train the morphological analyzer and the CCG parser to correctly handle a variety of temporal expressions.
We do not take this approach because it is too costly.

\input{sources/tree322_1_modified_tree.tex}

\subsection{Semantic Analysis}\label{subsec:semanticanalysis}
In semantic analysis, each leaf (lexical item) of the CCG tree obtained in the syntactic analysis is assigned a meaning from the semantic templates.
The lexical items are then combined according to the CCG derivation tree to derive a logical formula that expresses the meaning of the entire sentence.
The composition is performed using ccg2lambda in Japanese~\cite{mineshima-etal-2016-building}.

In order to assign meaning to the temporal expressions, we set up semantic templates for lexical items such as absolute temporal expressions and \textit{izen}.
We provide a set of semantic templates, which contains 150 lexical entries.
The number of lexical entries assigned to CCG categories is 92, and the number of entries directly assigned to specific words is 58.
Table~\ref{tab:semantic_templates} shows the examples of semantic templates.

As a representation language, we use the typed first-order form of the Thousands of Problems for Theorem Provers (TPTP; \citet{sutcliffe2017tptp}) format.
We use standard interval semantics~\cite{dowty1979word, bennett1978toward} and introduce an interval type to express time instances as intervals and their relations in logical expressions.
We use four basic types: $\mathsf{E}$ (Entity), $\mathsf{Ev}$ (Event), $\mathsf{Prop}$ (Proposition) and $\mathsf{I}$ (Interval).
The types of expressions we adopt are defined by
\begin{align*}
    \mathsf{T} ::= \mathsf{E}\ |\ \mathsf{Ev}\ |\ \mathsf{Prop}\ |\ \mathsf{I}\ |\ \mathsf{T1} \Rightarrow \mathsf{T2}
\end{align*}
where $\mathsf{T1} \Rightarrow \mathsf{T2}$ is a function type.
Because the logical expressions derived by ccg2lambda are not typed, we implement automatic completion of variable types, predicate types, and definitions of predicates.

\subsection{Theorem Proving}
In theorem proving, we use the state-of-the-art first-order logic automated theorem prover Vampire~\cite{kovacs2013first} which accepts TPTP formats to determine whether or not a hypothesis is provable from premises using the logical formula derived in Section~\ref{subsec:semanticanalysis}.
The system outputs ``yes'' (entailment) when the hypothesis can be proved from the premises, ``no'' (contradiction) when the negation of the hypothesis can be proved from the premises, and ``unknown'' (neutral) when neither can be proved.
We use the fastest mode, CASC mode, and set the timeout of Vampire to a maximum of 300 sec for our experiments.

Even though Vampire is a fast theorem prover, it takes too long to prove the problems, whose premises and hypothesis are too complex.
When proving the negation of a hypothesis, it turns out that simply negating the logical formula increases the complexity.
Therefore, this study uses the symmetrical relationship between \textit{ikou} and \textit{izen} to replace \textit{izen} and \textit{ikou} in the hypothesis with \textit{ikou} and \textit{izen}, respectively, to negate the logical formula without increasing the complexity.

\section{Experiments}
\subsection{Experimental Setup}\label{subsec:experimentalsetup}
We evaluate our system on two datasets.
First, JSeM~\cite{kawazoe2015inference} is a Japanese version of the FraCaS~\cite{cooper1996} test suite, which consists of nine sections, each containing representative problems of semantically challenging inferences involving various linguistic phenomena.
In this study, we use 23 problems involving temporal order in temporal reference section.
The distribution of gold answer labels for the problems is (yes/no/unknown) = (12/4/7).
\input{sources/table411_2_plmute_examples.tex}

\input{sources/table51_1_main_results.tex}
Second, we created an NLI dataset focusing on temporal order in Japanese from the existing NLI dataset (which we refer to as PLMUTE) for temporal inference in English proposed by \citet{thukral-etal-2021-probing} because Japanese NLI datasets involving diverse temporal adverbs were not well developed.
We used the ordering section of PLMUTE, which collects problems related to ordering various temporal adverbs for a date, day of the week, and time.
The original PLMUTE is automatically generated from 71 templates by a program.
Thus, we manually translated the templates into Japanese and modified the program to generate the dataset to make the generated dataset natural in Japanese.
We automatically generated a Japanese translation of the original PLMUTE by using the translated templates and modified program.
We call the dataset PLMUTE\_ja.
PLMUTE\_ja consists of nine sections: year (340 problems), month (480 problems), date (560 problems), date\_DMY (340 problems), date\_MY (340 problems), day (560 problems), time\_12 (400 problems), time\_24 (400 problems), and time\_multi (400 problems).
The distribution of gold answer labels for the problems is (yes/no/unknown) = (1353/1502/965).
Table~\ref{tab:plmute_examples} shows examples of problems in JSeM and PLMUTE\_ja.

We compared our system with the following previous logic-based inference system and deep learning-based models in Japanese.
\paragraph{Logic-based inference system}\label{subsubsec:exccg2lambda}
We used the logic-based inference system for temporal inference in Japanese proposed by \citet{Onishi2020ccg}.
\citet{Onishi2020ccg}'s system used Coq, a higher-order theorem prover based on natural deduction.

\paragraph{Deep learning-based model}\label{subsubsec:bert}
We used the Japanese BERT~\cite{devlin-etal-2019-bert} model (cl-tohoku/bert-base-japanese-whole-word-masking) of Huggingface transformers\footnote{\url{https://huggingface.co/transformers/}} as a deep learning-based model.
This Japanese BERT model is the most commonly used pre-trained language model for Japanese in huggingface/transformers.
In this study, we experimented with the following three models:
\textbf{BERT\_JSNLI} is Japanese BERT fine-tuned on a large Japanese NLI dataset JSNLI~\cite{yoshikoshi2020multilingualization} (533,005 examples), a Japanese translation of the SNLI dataset~\cite{bowman-etal-2015-large}, which is one of the most widely used NLI datasets.
\textbf{BERT\_few} is Japanese BERT fine-tuned on the PLMUTE\_ja minimal training set with two examples each of different combinations of tenses and sections (360 examples).
\textbf{BERT\_all} is Japanese BERT fine-tuned on the entire PLMUTE\_ja training set (11,220 examples).

\input{sources/table51_2_jsem_results.tex}
\section{Results and Discussion}\label{sec:resultsanddiscussion}

\subsection{Results}\label{subsec:results}
The results on the problems involving temporal order in JSeM are shown in Table~\ref{tab:jsem_results}.
As the table shows, our system outperforms all models.

The results on the PLMUTE\_ja test set are shown in Table~\ref{tab:main_results}.
As the table shows, our system outperforms all models except BERT\_all.
Although the performance is slightly inferior to BERT\_all, the performance is comparable to BERT\_all with 11,220 training data.
The experiment with Japanese BERT + PLMUTE\_ja reproduced the results of the experiment with English RoBERTa + PLMUTE conducted by \citet{thukral-etal-2021-probing}.
That is, although the model trained on all of the PLMUTE training sets could achieve high accuracy, the model trained on either the large standard NLI dataset or the minimal training set could only achieve low accuracy.

We also compared the average proof time for all four problems for which both our system and \citet{Onishi2020ccg}'s system output ``yes''.
Our system was faster than the previous logic-based system: the average proof time for our system was 1.98 seconds, while \citet{Onishi2020ccg}'s system was 3.11 seconds.

\subsection{Error Analysis}\label{subsec:erroranalysis}
In this section, we discuss the error analysis in the experiments.
Our system did not solve the problems involving comparative deletion and temporal connectives such as \textit{yori mae} (\textit{before}) and \textit{yori ato} (\textit{after}), as shown in Table \ref{tab:error_example}.

\begin{table}
    \centering
    \scalebox{0.78}{
    \begin{tabular}{ll}\bhline{1.5pt}
         $P_1$&ジョーンズが契約書を修正した。  \\
         &(Jones revised the contract.)\\
         $P_2$&スミスが契約書を修正した。   \\
         &(Smith revised the contract.)\\
         $P_3$&ジョーンズがスミスより先に契約書を修正した。   \\
         &(Jones revised the contract after Smith did.)\\
         $H$&スミスはジョーンズより後に契約書を修正した。\\
         &(Smith revised the contract before Jones did.)\\
         &\hspace{7.6em}Gold answer: yes (JSeM No. 659)\\ \bhline{1.5pt}
    \end{tabular}
    }
    \caption{An example of problem our system did not solve.}
    \label{tab:error_example}
\end{table}

\noindent Although \textit{yori mae} and \textit{yori ato} have similar meanings to \textit{izen} and \textit{ikou}, they have different meanings.
For example, 4 \textit{gatsu} 3 \textit{nichi izen} includes April 3rd, while 4 \textit{gatsu} 3 \textit{nichi yori-mae} does not include April 3rd.
In addition, \textit{yori mae} is more difficult to analyze than \textit{izen} because it consists of two words, \textit{yori} and \textit{mae} that require the analysis of comparative deletion, which we leaves for future work.

\section{Conclusion}
In this paper, we compositionally derived semantic representations of sentences with tense and aspect in Japanese based on CCG.
We developed a logic-based NLI system that considers temporal order in Japanese.
We evaluated our system by experimenting with two Japanese NLI datasets involving temporal order.
Our system performed more robustly than previous logic-based systems as well as current deep learning-based models.
The experimental results of our system suggest that a logical NLI system based on an analysis of tense in formal semantics is effective for temporal inference.
Other previous studies of logic-based methods have shown the effectiveness of NLI systems based on the analysis of various semantics such as degree semantics~\cite{haruta-etal-2020-logical}.
By combining them, we will be able to construct one NLI system capable of performing a variety of inferences.
In the future, we plan to cover various temporal inferences involving comparative deletion and temporal anaphora.
Furthermore, we plan to construct inference test sets for these challenging inferences.

\section*{Acknowledgements}
We thank the three anonymous reviewers for their helpful comments and feedback.
This work was supported by PRESTO, JST Grant Number JPMJPR21C8, Japan.

\bibliography{anthology,custom}
\bibliographystyle{acl_natbib}

\appendix

\end{document}

%% file: sources/table232_1_tense_aspect_relation.tex
\begin{table}[ht]
\centering
\scalebox{0.97}{
\begin{tabular}{llcl}
\bhline{1.5pt}
Past                 & Stative & Relation & Example                                                              \\ \hline\hline
\multirow{2}{*}{$+$} & $+$  & $r < s$   & \begin{tabular}[c]{@{}l@{}}\textit{Taro-ga ita}\\ ‘Taro was here’\end{tabular}   \\
                     & $-$  & $r < s$   & \begin{tabular}[c]{@{}l@{}}\textit{Taro-ga kita}\\ ‘Taro came’\end{tabular}      \\
\multirow{2}{*}{$-$} & $+$  & $r \ge s$ & \begin{tabular}[c]{@{}l@{}}\textit{Taro-ga iru}\\ ‘Taro is here’\end{tabular}    \\
                     & $-$  & $r > s$   & \begin{tabular}[c]{@{}l@{}}\textit{Taro-ga kuru}\\ ‘Taro is coming’\end{tabular} \\ \bhline{1.5pt}
\end{tabular}
}
\caption{Constraints imposed on the relation between speech time $s$ and reference time $r$ by tense and aspect and example sentences}
\label{tab:tense_aspect_relation}
\end{table}

%% file: sources/tree311_1_example_derivation.tex
\begin{figure*}[t]
\centering
\tiny

~\deduce{\SR{\exists s r.(\top \,\wedge\, \exists e1.(\mathsf{come}(e1) \,\wedge\, \mathsf{during}(\mathsf{time}(e1),r) \,\wedge\, \mathsf{after}(r,s) \,\wedge\, (\mathsf{Nom}(e1) = \mathsf{Taro})))}}{
\infer[]{S_{[nm,base,t]}}{
\deduce{\SR{\lambda C1 C2 C3 K i1 j1.(\top \,\wedge\, \exists e1.(K(\lambda e2 i2 j2.(\mathsf{come}(e2) \,\wedge\, \mathsf{during}(\mathsf{time}(e2),j2) \,\wedge\, \mathsf{after}(j2,i2)),e1,i1,j1) \,\wedge\, C1(\mathsf{Taro},e1,\mathsf{Nom})))}}{
\infer[\FA]{S_{[nm,base,f]}}{
\deduce{\SR{\lambda N F.(N(\lambda x.\top,\mathsf{Taro}) \,\wedge\, F(\mathsf{Taro}))}}{
\infer[\FA]{NP_{[ga,nm,f]}}{
\deduce{\SR{\lambda N F.(N(\lambda x.\top,\mathsf{Taro}) \,\wedge\, F(\mathsf{Taro}))}}{
\infer{NP_{[nc,nm,f]}}{\mbox{太郎 (Taro)}}} &
\deduce{\SR{\lambda Q.Q}}{
\infer{NP_{[ga,nm,f]}\backslash NP_{[nc,nm,f]}}{\mbox{が (NOM)}}}}} &
\deduce{\SR{\lambda Q C1 C2 C3 K i1 j1.Q(\lambda I.I,\lambda x.\exists e1.(K(\lambda e2 i2 j2.(\mathsf{come}(e2) $\\$\,\wedge\, \mathsf{during}(\mathsf{time}(e2),j2) \,\wedge\, \mathsf{after}(j2,i2)),e1,i1,j1) \,\wedge\, C1(x,e1,\mathsf{Nom})))}}{
\infer{S_{[nm,base,f]}\backslash NP_{[ga,nm,f]}}{\mbox{来る (come-NP)}}}}}}}

\caption{CCG derivation tree for \textit{Taro-ga kuru} (\textit{Taro is coming}). $\top$ denotes the tautology.}
\label{fig:example_derivation}
\end{figure*}

%% file: sources/tree321_1_example_CCG_tree.tex
\begin{figure}[ht]
\small
\centering
~\infer{S/S}{
{
\infer{S/S}{
{
\infer{NP}{\mbox{4月3日 (April 3)}}} &
{
\infer{(S/S)\backslash NP}{\mbox{以前 (before)}}}}} &
{
\infer{(S/S)\backslash (S/S)}{\mbox{に}}}}

\caption{CCG derivation tree for 4 \textit{gatsu} 3 \textit{nichi izen ni} (\textit{before April} 3).}
\label{fig:example_ccg_tree}
\end{figure}

%% file: sources/table332_1_semantic_templates.tex
\begin{table*}[t]
\centering
\small
\begin{tabular}{lll}
\bhline{1.5pt}
Category                            & Expression                                                 & Semantic Template                                                                                                                                                                                                                                                                           \\ \hline\hline
$S\backslash NP$       & \begin{tabular}[c]{@{}l@{}}来る\\ (is coming)\end{tabular}   & \begin{tabular}[c]{@{}l@{}}$\lambda Q\ C1\ C2\ C3\ K\ i1\ j1.Q(\lambda I.I,\lambda x.^\exists e1.(K(\lambda e2\ i2\ j2.(\mathsf{come}(e2) $\\ $\,\wedge\, \mathsf{during}(\mathsf{time}(e2),j2)\,\wedge\, \mathsf{after}(j2,i2)),e1,i1,j1) \,\wedge\, C1(x,e1,\mathsf{Nom})))$\end{tabular}  \\ \hline
$S\backslash NP$       & \begin{tabular}[c]{@{}l@{}}来た\\ (came)\end{tabular}        & \begin{tabular}[c]{@{}l@{}}$\lambda Q\ C1\ C2\ C3\ K\ i1\ j1.Q(\lambda I.I,\lambda x.^\exists e1.(K(\lambda e2\ i2\ j2.(\mathsf{come}(e2) $\\ $\,\wedge\, \mathsf{during}(\mathsf{time}(e2),j2)\,\wedge\, \mathsf{before}(j2,i2)),e1,i1,j1) \,\wedge\, C1(x,e1,\mathsf{Nom})))$\end{tabular} \\ \hline
$NP$                                  & \begin{tabular}[c]{@{}l@{}}4月3日\\ (April 3rd)\end{tabular} & $\lambda N\ F.^\exists x.(N(\lambda y.(\mathsf{normalized\_time}(y) = 40300),x)\,\wedge\, F(x))$                                                                                                                                                                                            \\ \hline
$S/S$                                 & \begin{tabular}[c]{@{}l@{}}4月3日\\ (April 3rd)\end{tabular} & \begin{tabular}[c]{@{}l@{}}$\lambda S\ C1\ C2\ C3\ K\ i1\ j1.S(C1,C2,C3,\lambda J\ e1\ i2\ j2.K(\lambda e2\ i3\ j3.(J(e2,i3,j3)\,\wedge\,$\\ $^\exists x.((\mathsf{normalized\_time}(x) = 40300)\,\wedge\,(x = j3))),e1,i2,j2),i1,j1)$\end{tabular}                                                         \\ \hline
$(S/S)\backslash NP$   & \begin{tabular}[c]{@{}l@{}}以前\\ (before)\end{tabular}      & \begin{tabular}[c]{@{}l@{}}$\lambda Q\ S\ C1\ C2\ C3\ K\ i1\ j1. S(C1,C2,C3,\lambda J\ e1\ i2\ j2.K(\lambda e2\ i3\ j3.(J(e2,i3,j3)\,\wedge\,$\\ $Q(\lambda I.I, \lambda x.\mathsf{before}(j3,x))),e1,i2,j2),i1,j1)$\end{tabular}                                                           \\ \hline
$(S/S)\backslash(S/S)$ & \begin{tabular}[c]{@{}l@{}}に\\ (on)\end{tabular}     & $\lambda V3. V3$                                                                                                                                                                                                                                                                            \\ \bhline{1.5pt}
\end{tabular}
\caption{Examples of semantic templates.}
\label{tab:semantic_templates}
\end{table*}

%% file: sources/table341_1_axioms.tex
\begin{table*}[t]
\centering
\scalebox{0.8}{
\begin{tabular}{ll}
\bhline{1.5pt}
Pattern                                                           & \multicolumn{1}{l}{Axiom}                                                                                                                                                                                                                                                                                \\ \hline\hline
transitivity of $\mathsf{before}$ relations                                                    & $^\forall A, B, C. (\mathsf{before}(A, B) \wedge \mathsf{before}(B, C) \rightarrow \mathsf{before}(A, C))$                               \\ \hline
insertion of \textit{izen}                                                           & \begin{tabular}[c]{@{}l@{}}$^\forall I,X,R. ((\mathsf{nort}(X) = I \wedge (X = R)) \rightarrow (^\forall J. ((I\le J) $\\ $\rightarrow (^\exists Y. (\mathsf{nort}(Y) = J \wedge ^\exists Z. (\mathsf{before}(Z, Y) \wedge (Z = R))))))).$\end{tabular}                                                  \\ \hline
replacement of \textit{izen} & \begin{tabular}[c]{@{}l@{}}$^\forall I,X,R. ((\mathsf{nort}(X) = I \wedge \mathsf{before}(R, X)) \rightarrow (^\forall J. ((I\le J)$\\ $\rightarrow (^\exists Z. (\mathsf{nort}(Z) = J \wedge \mathsf{before}(R, Z))))))$\end{tabular} \\ \hline
\begin{tabular}[c]{@{}l@{}}identity condition of speech times\end{tabular} & $^\forall S_1,S_2. (\mathsf{speech\_time}(S_1) \wedge \mathsf{speech\_time}(S_2) \rightarrow S_1 = S_2)$                                                                                                                                                                                                        \\ \bhline{1.5pt}
\end{tabular}
}
\caption{Examples of axioms. $\mathsf{nort}$ indicates a $\mathsf{normalized\_time}$ function.}
\label{tab:axioms}
\end{table*}

%% file: sources/tree322_1_modified_tree.tex
\begin{figure}[ht]
\tiny
\centering

~\infer[\FA]{S/S}{
{
\infer[\BA]{NP}{
{
\infer[\BA B]{NP/NP}{
{
\infer{NP/NP}{\mbox{4}}} &
{
\infer{NP/NP}{\mbox{月(month)}}}}} &
{
\infer[\FA]{NP}{
{
\infer{NP}{\mbox{3}}} &
{
\infer{NP\backslash NP}{\mbox{日(day)}}}}}}} &
{
\infer{(S/S)\backslash NP}{\mbox{に}}}}
\captionof{figure}{CCG derivation tree before conversion.}
\label{fig:before_modify_tree}
\end{figure}

\begin{figure}[ht]

\small
\centering
~\infer[\FA]{S/S}{
{
\infer{S/S}{\mbox{4月3日(April 3)}}} &
{
\infer{(S/S)\backslash (S/S)}{\mbox{に}}}}

\captionof{figure}{CCG derivation tree after conversion.}
\label{fig:after_modify_tree}

\vspace{-5mm}
\end{figure}

%% file: sources/table411_2_plmute_examples.tex
\begin{table}[ht]
\centering
\small
\begin{tabular}{ll}
\bhline{1.5pt}
\multicolumn{2}{l}{PLMUTE Section: time\_multi, No. 11, Gold answer: yes}                                                              \\ \hline
$P$    & \begin{tabular}[c]{@{}l@{}}午後7時以降ロビンは両親を訪ねた。\\ (After 7 p.m. Robin visited her parents.)\end{tabular}     \\
$H$ & \begin{tabular}[c]{@{}l@{}}16時以降ロビンは両親を訪ねた。\\ (After 16:00 Robin went to visit her parents.)\end{tabular} \\ \hline
\multicolumn{2}{l}{PLMUTE Section: day, No. 239, Gold answer: no}                                                                               \\ \hline
$P$    & \begin{tabular}[c]{@{}l@{}}月曜日以前、食料品店が閉店した。\\ (Before Monday, the grocery store was closed.)\end{tabular} \\
$H$ & \begin{tabular}[c]{@{}l@{}}火曜日以降、食料品店が閉店した。\\ (After Tuesday, the grocery store was closed.)\end{tabular} \\ \hline\hline
\multicolumn{2}{l}{JSeM No. 645, Gold answer: yes}                                                                               \\ \hline
$P$    & \begin{tabular}[c]{@{}l@{}}1992年以来、ITELはバーミンガムにある。 \\
(Since 1992 ITEL has been in Birmingham.)\\
現在、1996年である。 \\
(It is now 1996.)\end{tabular} \\
$H$ & \begin{tabular}[c]{@{}l@{}}
ITELは1993年にはバーミンガムにあった。\\ (ITEL was in Birmingham in 1993.)\end{tabular} \\ \bhline{1.5pt}
\end{tabular}
\caption{Examples of problems from JSeM and PLMUTE\_ja.}
\label{tab:plmute_examples}
\end{table}

%% file: sources/table51_1_main_results.tex
\begin{table*}[ht]
\centering
\begin{tabular}{ll|lllllllll}
\bhline{1.5pt}
\multicolumn{2}{l|}{System}                       & year  & month & date & \begin{tabular}[c]{@{}l@{}}date\\ \_dmy\end{tabular} & \begin{tabular}[c]{@{}l@{}}date\\ \_my\end{tabular} & day   & \begin{tabular}[c]{@{}l@{}}time\\ \_12\end{tabular} & \begin{tabular}[c]{@{}l@{}}time\\ \_24\end{tabular} & \begin{tabular}[c]{@{}l@{}}time\\ \_multi\end{tabular} \\ \hline \hline
\multicolumn{2}{l|}{Majority}                      & .382  & .421  & .425 & .403                                                 & .379                                                & .396  & .368                                                & .415                                                & .418                                                   \\ \hline
\multicolumn{1}{l|}{\multirow{3}{*}{BERT}} & JSNLI & .394  & .413  & .382 & .400                                                 & .400                                                & .380  & .378                                                & .415                                                & .368                                                   \\ \cline{2-2}
\multicolumn{1}{l|}{}                      & few   & .509  & .517  & .509 & .491                                                 & .476                                                & .518  & .440                                                & .453                                                & .515                                                   \\ \cline{2-2}
\multicolumn{1}{l|}{}                      & all   & .997  & 1.000  & .998 & .985                                                 & .982                                                & 1.000 & 1.000                                                & .998                                                & .960                                                   \\ \hline
\multicolumn{2}{l|}{Onishi et al. (2020)}          & .238  & .265  & .239 & .206                                                 & .244                                                & .291  & .290                                                & .225                                                & .253                                                   \\ \hline
\multicolumn{2}{l|}{\textbf{Our system}}           & 1.000 & 1.000  & .980 & .971                                                 & .974                                                & .984  & .943                                                & .970                                                & .953                                                   \\ \bhline{1.5pt}
\end{tabular}
\caption{Accuracy on the PLMUTE\_ja test suite.}
\label{tab:main_results}
\end{table*}

%% file: sources/table51_2_jsem_results.tex
\begin{table}[t]
\centering
\begin{tabular}{llc}
\bhline{1.5pt}
\multicolumn{2}{l|}{System}&Accuracy                                                                 \\ \hline\hline
\multicolumn{1}{l|}{\multirow{3}{*}{BERT}} & \multicolumn{1}{l|}{JSNLI} & .522           \\
\multicolumn{1}{l|}{}                      & \multicolumn{1}{l|}{few}   & .217           \\
\multicolumn{1}{l|}{}                      & \multicolumn{1}{l|}{all}   & .435           \\  \hline
\multicolumn{2}{l|}{Onishi et al. (2020)}                                & .478           \\ \hline
\multicolumn{2}{l|}{\textbf{Our system}}                                & .783           \\ \bhline{1.5pt}
\end{tabular}
\caption{Accuracy on the problems involving temporal order in the JSeM test suite.
}
\label{tab:jsem_results}
\end{table}